\documentclass[journal=jacsat,manuscript=article]{achemso}

\usepackage[version=3]{mhchem} 
\usepackage[noabbrev]{cleveref}
\usepackage{xspace}
\usepackage{algorithm}
\usepackage{algpseudocode}
\usepackage{graphicx}
\usepackage{caption}
\usepackage{subcaption}
\usepackage{multirow}
\usepackage{makecell}
\usepackage{booktabs}
\usepackage{xcolor}
\usepackage{enumerate}
\usepackage{xurl}
\usepackage{threeparttable}
\usepackage{rotating}
\usepackage{longtable}
\usepackage{bbm}

\newcommand{\ours}{RxnScribe\xspace}
\newcommand{\rev}[1]{{#1}}

\author{Yujie Qian}
\email{yujieq@csail.mit.edu}
\author{Jiang Guo}
\author{Zhengkai Tu}
\affiliation{Computer Science and Artificial Intelligence Laboratory, MIT, Cambridge, MA, 02139}
\author{Connor W. Coley}
\affiliation{Department of Chemical Engineering, MIT, Cambridge, MA, 02139}
\author{Regina Barzilay}
\affiliation{Computer Science and Artificial Intelligence Laboratory, MIT, Cambridge, MA, 02139}
\email{regina@csail.mit.edu}

\title[RxnScribe]{RxnScribe: A Sequence Generation Model for Reaction Diagram Parsing}

\begin{document}

\begin{abstract}
Reaction diagram parsing is the task of extracting reaction schemes from a diagram in the chemistry literature. The reaction diagrams can be arbitrarily complex, thus robustly parsing them into structured data is an open challenge. In this paper, we present \ours, a machine learning model for parsing reaction diagrams of varying styles. We formulate this structured prediction task with a sequence generation approach, which condenses the traditional pipeline into an end-to-end model. We train \ours on a dataset of 1,378 diagrams and evaluate it with cross validation, achieving an 80.0\% soft match F1 score, with significant improvements over previous models. Our code and data are publicly available at \url{https://github.com/thomas0809/RxnScribe}.
\end{abstract}

\section{Introduction}

\begin{figure}[t!]
    \centering
    \includegraphics[width=\linewidth]{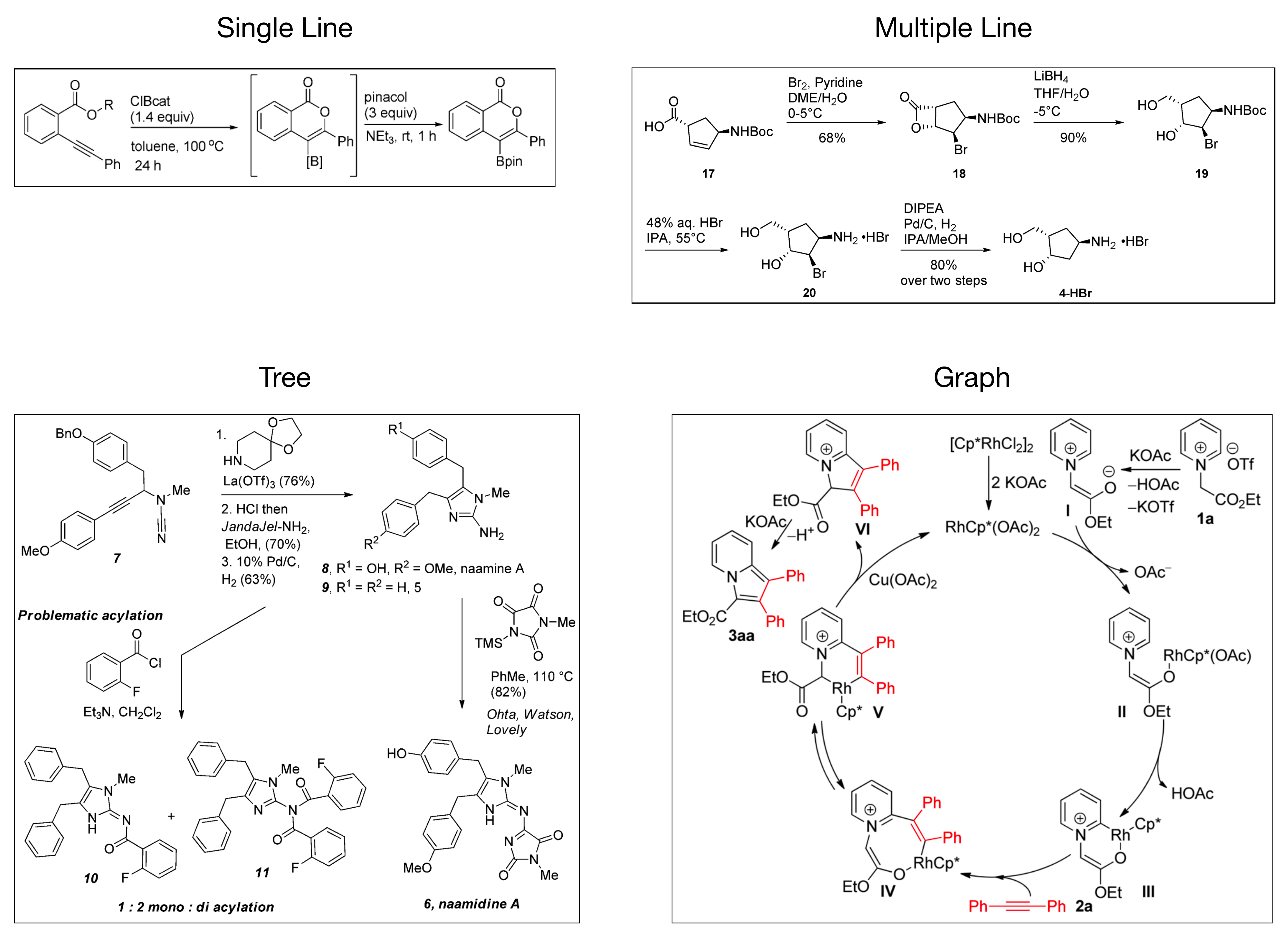}
    \caption{Examples of reaction diagrams in chemistry literature. We summarize four common styles of reaction diagrams: single-line, multiple-line, tree, and graph. The example diagrams are adapted with permission from 
    \citet{faizi2016catalyst} [Copyright © 2016 American Chemical Society], \citet{armitage2015process} [Copyright © 2015 American Chemical Society],
    \citet{gibbons2015synthesis} [Copyright © 2015 American Chemical Society],
    and
    \citet{shen2016rh} [Copyright © 2016 American Chemical Society].  }
    \label{fig:diagram}
\end{figure}

In the chemistry literature, new reactions and synthesis pathways are often presented in diagrams. As \Cref{fig:diagram} illustrates, these diagrams exhibit significant diversity and can be arbitrarily complex. The importance of automatic parsing of these diagrams into structured data has been recognized by the research community.\cite{mse-staker,csr,rde} For each diagram, the task is to recognize the reaction scheme and extract the reactants, conditions, and products for each reaction. 
In this paper, we aim to design a general machine learning solution for  reaction diagram parsing that robustly generalizes across styles. Assuming expert annotations of reaction schemes on a collection of diagrams, we train a neural network model that can predict the reactions in new diagrams.

This paper introduces \ours, a simple and effective model for reaction diagram parsing. We formulate this structured prediction problem as sequence generation. To accomplish this, we define a sequence representation to describe the reaction structure in a diagram, which specifies the reaction roles (reactants, conditions, and products) and indicates the bounding box coordinates and the type of each entity (e.g., a molecular graph or a textual description). We train a generation model to predict this sequence representation conditioned on the diagram image. Compared to traditional pipelined approaches\cite{rde} that first extract the entities and then predict their relationships, our formulation simplifies the process and naturally avoids the problem of error propagation. At inference time, we decode the reaction structure from the predicted sequence, and subsequently apply off-the-shelf molecular structure recognition \cite{molscribe} and optical character recognition \cite{easyocr} models to translate the entity bounding boxes into molecular structures and texts.

To train \ours, we construct a dataset with reaction diagrams collected from the chemistry literature. The dataset consists of 1,378 diagrams with 3,776 reactions, covering the four styles presented in \Cref{fig:diagram}. The ground truth of reaction structure is annotated by domain experts. 
We further develop a data augmentation strategy that composes simple diagrams into more complex ones to augment the training data. 

In the experiments, we evaluate \ours on our dataset with five-fold cross validation. We only evaluate the accuracy of the predicted reaction structure, as we do not have the ground truth for the molecular structures (i.e., SMILES strings) and text content. \ours attains a significant performance boost compared to previous models. The model achieves an 80.0\% F1 score (soft match) overall, ranging from 91.0\% on single-line diagrams to 65.9\% on the most complicated graph-style diagrams, while the scores for existing models are below 10\%. \ours also benefits from the proposed data augmentation techniques to achieve strong performance with a small number of training data. Our code and data are publicly available at \url{https://github.com/thomas0809/RxnScribe}.  We have also developed an online interface for \ours: \url{https://huggingface.co/spaces/yujieq/RxnScribe}.

\section{Reaction Diagram Parsing}
\begin{figure}[t!]
    \centering
    \includegraphics[width=\linewidth]{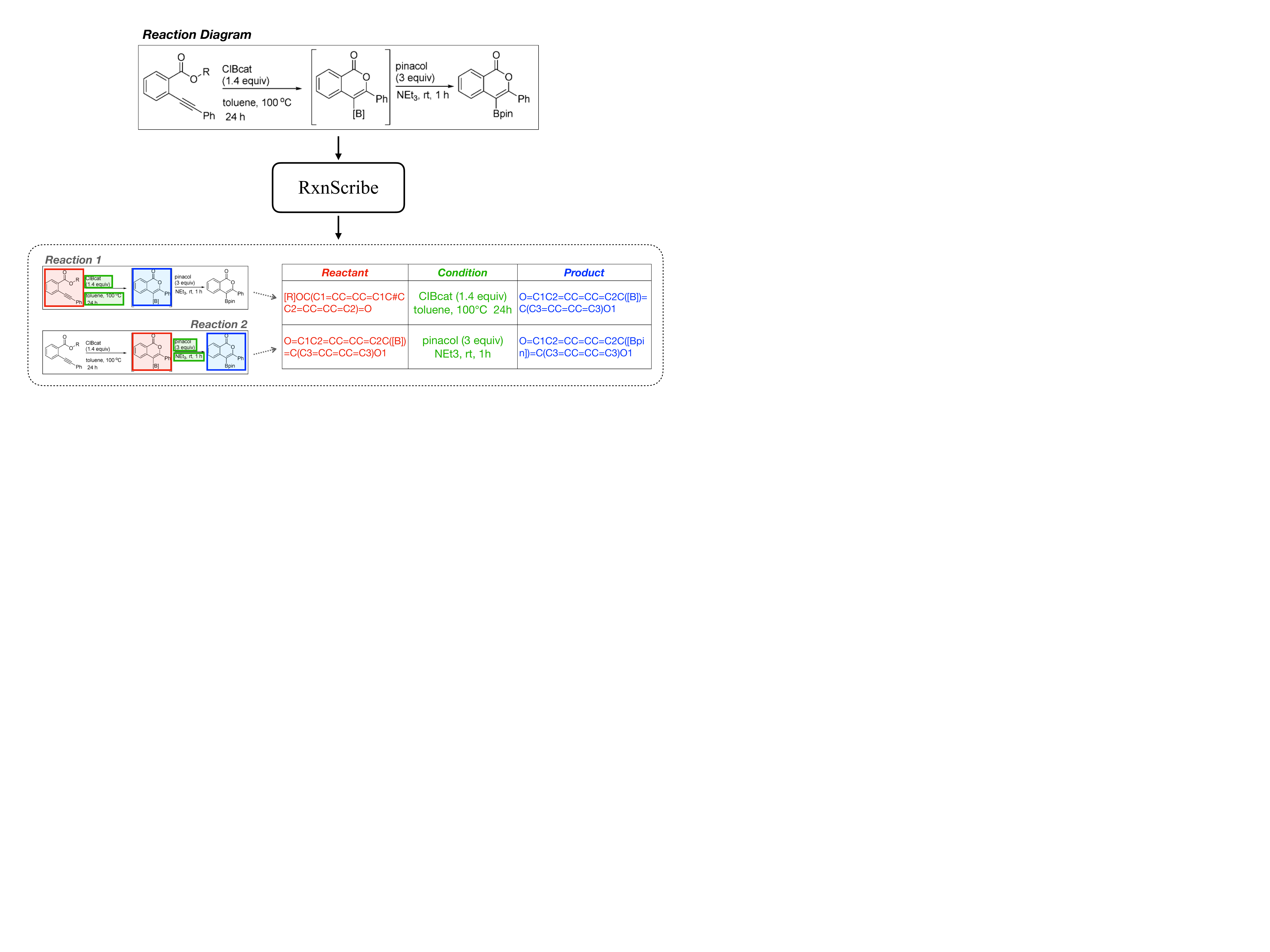}
    \caption{Overview of reaction diagram parsing. The input is a reaction diagram, and the output is a list of reactions. The example diagram is from a journal article.\cite{faizi2016catalyst}}
    \label{fig:model}
\end{figure}

\subsection{Task Definition}
Reaction diagram parsing is the task of extracting chemical reactions from the diagrams in the literature. \Cref{fig:model} gives an overview of the task. The input is an image of a reaction diagram, which illustrates either a single reaction or a series of reactions. We aim to extract the reaction(s) in this diagram, and identify their reactants, products, and reaction conditions. Specifically, the expected output is a list of reactions $\{R_1, R_2,\dots, R_n\}$, where each reaction consists of three roles $R_i = (S_i, C_i, T_i)$. $S_i$ is the set of reactants  and $T_i$ is the set of products, each consisting of one or multiple molecule structures. $C_i$ is the set of reaction conditions, which may be empty if no condition is specified in the diagram.

To obtain the input for our task, we utilize PDF parsing tools\cite{pdffigures} to crop reaction diagrams from chemistry papers and convert them into image files (PNG format).

\subsection{Related Work}
Published research on extracting reactions from chemistry literature focuses primarily on text.\cite{oscar4,chemicaltagger,lowe2012extraction,cde} For example, Pistachio\cite{pistachio} is a reaction dataset constructed from patent text with a traditional natural language processing pipeline, which includes syntactic parsing, named entity recognition (to extract chemical names), and event extraction (to assemble chemicals into reactions).\cite{lowe2020extraction,ChEMU} \citeauthor{science.aav2211} and \citeauthor{vaucher2020automated} developed expert-curated heuristics\cite{science.aav2211} and a sequence-to-sequence generation model\cite{vaucher2020automated}, respectively, to convert experimental procedure text into synthesis actions. 
For processing more diverse text in journal articles, \citeauthor{guo2021automated} proposed a deep learning model to extract the reaction schemes.\cite{guo2021automated} They formulated the task into two stages: product extraction and reaction role labeling, each solved by sequence tagging adapted from pre-trained language models. Our paper studies a different input source -- reaction diagrams, which are images and naturally require different models to process. Both text and diagram parsing are important components in information extraction from chemistry literature.

Prior work on diagram parsing focused on the segmentation of molecular images from the diagrams\cite{mse-staker,csr} and the recognition of their chemical structures\cite{osra,mse-staker,decimer,chemgrapher,molscribe}. 
Only a few attempted to understand the relationships between the molecules, i.e., reaction schemes. 
\citeauthor{rde} proposed ReactionDataExtractor\cite{rde} to extract reaction schemes from the diagrams. They developed a pipeline based on image processing techniques and heuristics. This method first converts the input image to grayscale and removes noisy pixels. Then, additional rules are used to segment out the arrows, molecules, and texts. For example, arrows are identified by running a line detection algorithm, and filtered based on a criterion that the half with the arrow hook should have sufficiently more pixels than the other half. Molecules are identified by first clustering the pixels, then finding the clusters with many bonds, suitable sizes, and aspect ratios. Finally, they use an arrow as the indicator of a reaction and assign reactants, products, and conditions to the arrow according to their relative positions and distances. ReactionDataExtractor can successfully parse simple single-line reaction diagrams, but many diagrams in chemistry literature contain patterns that are not covered by this rule-based system. For example, reactions presented with vertical, branched, or curved arrows (see \Cref{fig:diagram}) cannot be recognized by their system. 
In our experiments, we found that the performance of those heuristics on our collected diagrams is unsatisfactory given realistic variation in drawing styles. 


\subsection{Model}

We propose a general-purpose neural network framework \ours for reaction diagram parsing. \ours is an extraction model that identifies the reaction structure in the image and segments out the relevant entities, i.e., their reactants, conditions, and products. Then, we use MolScribe \cite{molscribe}, a molecular structure recognition model, to translate the images of molecular entities into SMILES strings, and use an optical character recognition (OCR) tool \cite{easyocr} to recognize the text content. In this paper, we mainly discuss the reaction extraction model. 

\begin{figure}[t!]
    \centering
    \includegraphics[width=\linewidth]{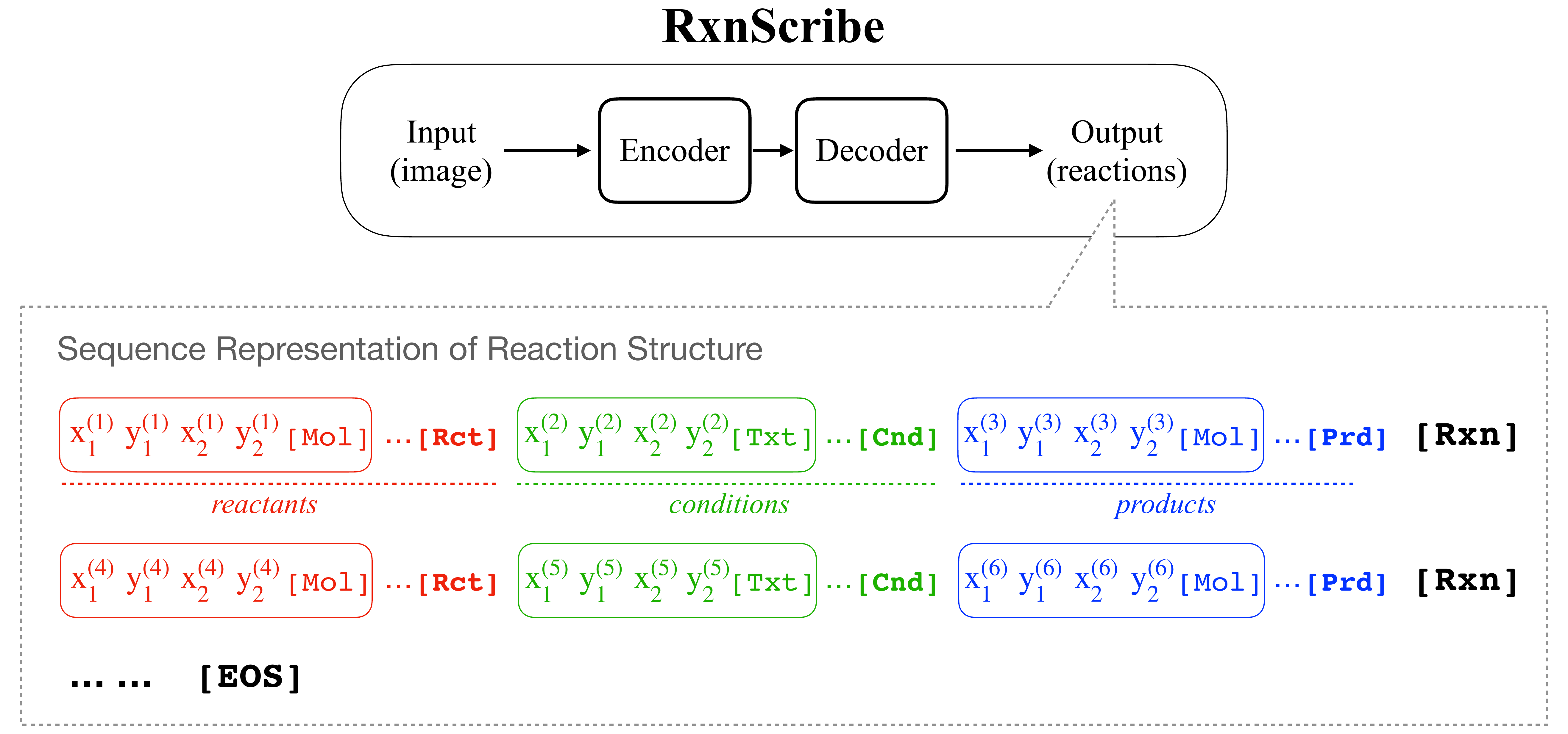}
    \caption{\ours is a sequence generation model for reaction diagram parsing. We define a sequence representation of the reaction structure in a diagram. Each entity is represented as five tokens. The reaction role is described with special tokens (\texttt{[Rct]}: reactant, \texttt{[Cnd]}: condition, \texttt{[Prd]}: product, \texttt{[Rxn]}: reaction). }
    \label{fig:seq}
\end{figure}

The challenge of formulating reaction diagram parsing with machine learning lies in the complexity of the reaction structure. Each reaction role may contain a variable number of entities (molecules or texts), thus we cannot apply existing information extraction models which usually predict the relationship between two entities\rev{\cite{MiwaB16,ZhangZCAM17,LuanHOH18}}.
In this paper, we propose a simple and effective formulation. We define a sequence representation of the reaction structure in a diagram, which serializes the reactions into a sequence of tokens, and train a model to generate this sequence given the diagram.

\paragraph{Sequence Representation of Reaction Structure}

\Cref{fig:seq} illustrates the definition of our sequence representation of reaction structure in a diagram. First, each entity of interest is represented as five tokens: 
\begin{equation}
\textsc{Entity}\ :=\ \mathrm{x_1} \ \mathrm{y_1} \ \mathrm{x_2} \ \mathrm{y_2} \ \textsc{EntityType} 
\end{equation} 
The first four tokens describes its bounding box in the image, $(\mathrm{x_1}, \mathrm{y_1})$ and $(\mathrm{x_2}, \mathrm{y_2})$ are the coordinates of the top-left and bottom-right corners, respectively. The coordinates are converted to integer tokens by binning\cite{pix2seq}, i.e.,
$\mathrm{x} := \lfloor \frac{x}{W} \times n_\text{bins} \rfloor,\   \mathrm{y} := \lfloor \frac{y}{H} \times n_\text{bins} \rfloor$,
where $x,y$ are the pixel-level coordinates, $W$ and $H$ are the width and height of the diagram image, and $n_\text{bins}$ is a hyperparameter for the number of bins.
The fifth token represents the entity type:
\begin{equation}
\textsc{EntityType}\ :=\ \texttt{[Mol]} \mid \texttt{[Txt]} \mid \texttt{[Idt]}
\end{equation}
We define three types of entities: molecule (\texttt{[Mol]}), text (\texttt{[Txt]}), and identifier (\texttt{[Idt]}). (Identifier is a text label which refers to another molecule in the same article, such as \textbf{2} and \textbf{3c}.) Usually, reactants and products are drawn as molecular graphs and conditions are written in text in the diagram, but there are also many cases where molecules are denoted as text or identifiers.

The sequence for the reaction structure in a diagram is defined by the following grammar:
\begin{alignat}{2}
&\textsc{ReactionStructure} &&:=\ (\textsc{Reaction})^* \ \texttt{[EOS]} \\
&\textsc{Reaction} &&:=\ \textsc{Reactants}\quad \textsc{Conditions}\quad \textsc{Products}\quad \texttt{[Rxn]} \\
&\textsc{Reactants} &&:=\ (\textsc{Entity})^+ \ \texttt{[Rct]} \\
&\textsc{Conditions} &&:=\ (\textsc{Entity})^* \ \texttt{[Cnd]} \\
&\textsc{Products} &&:=\ (\textsc{Entity})^+ \ \texttt{[Prd]}
\end{alignat}
where $(\cdot)^*$ means zero or more occurrences, and $(\cdot)^+$ means one or more occurrences.
Each $\textsc{Reaction}$ is a subsequence, which consists of three reaction roles: $\textsc{Reactants}$, $\textsc{Conditions}$, and $\textsc{Products}$, and ends with a $\texttt{[Rxn]}$ token. Each reaction role corresponds to a sequence of \textsc{Entity}s, and ends with a special token (\texttt{[Rct]}, \texttt{[Cnd]}, or \texttt{[Prd]}). Note that \textsc{Reactants} and \textsc{Products} must contain at least one \textsc{Entity}, but \textsc{Conditions} can be empty. The overall $\textsc{ReactionStructure}$ is described by stacking the \textsc{Reaction} subsequences one by one. We rank the \textsc{Reaction}s according to their reading order (explained in the data section). Finally, an \texttt{[EOS]} token completes the sequence. 


\paragraph{Model Architecture}
RxnScribe takes a diagram image as input and generates a sequence of the reaction structure. The model has an encoder-decoder architecture: an encoder abstracts the input image into hidden representations, and a decoder generates the output sequence in an autoregressive fashion, i.e., it predicts one token at a time, conditioned on the image encoding and the tokens that has already been generated. We follow the implementation of Pix2Seq\cite{pix2seq}, which was originally designed for object detection. It uses a convolutional neural network as the encoder, and a Transformer network as the decoder.

\paragraph{Training}
\ours is trained by maximizing the likelihood of the ground truth reaction structure sequence via teacher forcing. 
In practice, we first pre-train the model on a generic object detection dataset\rev{\cite{mscoco}}, which only seeks to predict the objects (entities) in an image, then finetune the model on our reaction diagram dataset. The pre-training helps the model to converge faster and improves its final performance.

\begin{figure}[t]
    \centering
    \includegraphics[width=0.85\linewidth]{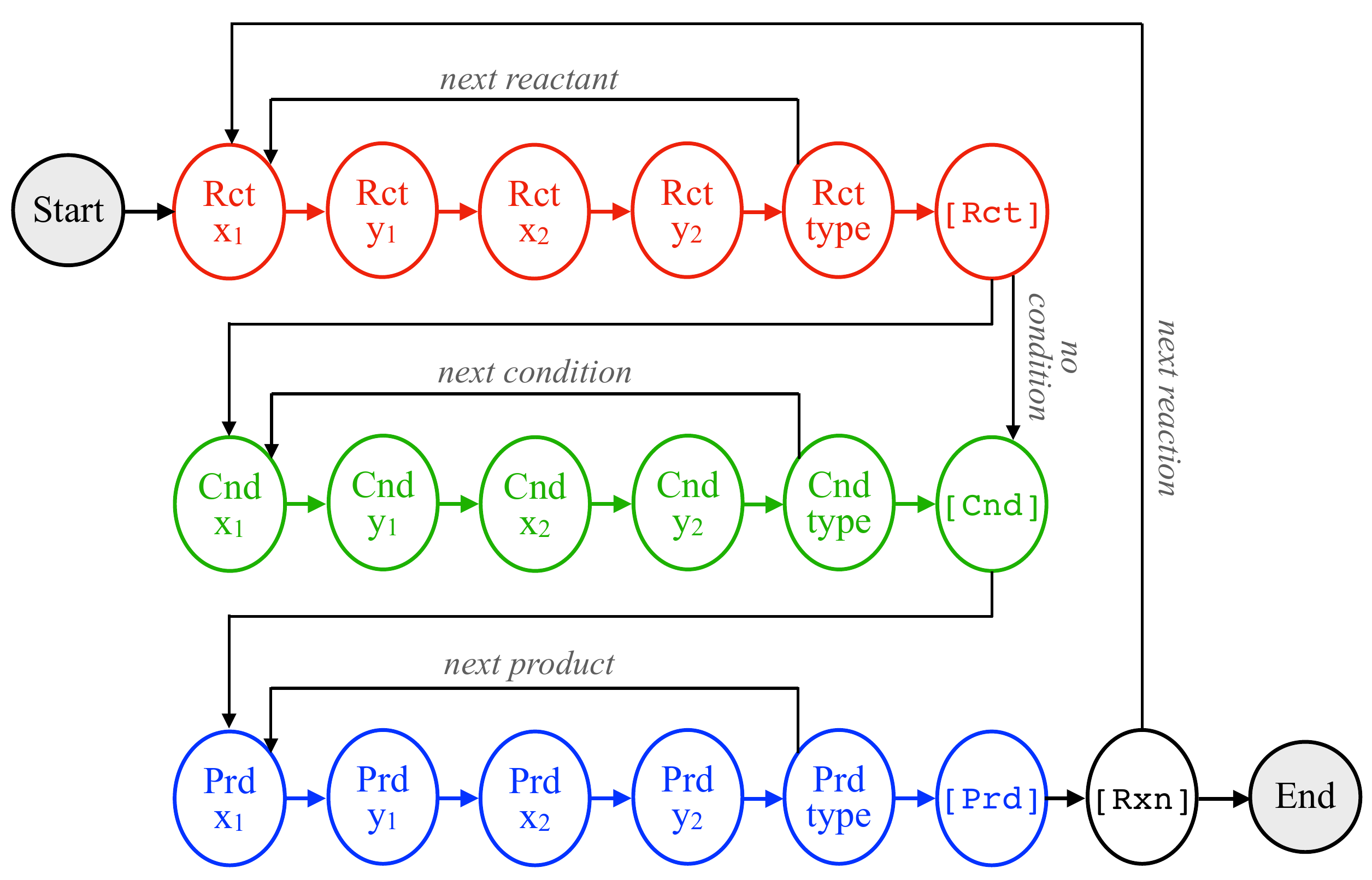}
    \caption{Transition between states at inference time. Circles are the states and arrows are the possible transitions. Each token generated is mapped to a state in this figure.}
    \label{fig:inference}
\end{figure}

\paragraph{Inference}
At inference time, \ours uses a greedy decoding strategy to generate its output sequence. We impose simple constraints to guarantee the generation follows the grammar of the proposed sequence representation. When \ours is decoding, we maintain a \textit{state} of the current prediction step, and use it to determine what tokens the model is allowed to predict in the next step. \Cref{fig:inference} displays the possible transitions between the states. For example, when the model is predicting the $\mathrm{x}_1$ coordinate of a reactant (state ``Rct $\mathrm{x}_1$''), it must predict its $\mathrm{y}_1$ coordinate next; when the model is predicting the entity type of a product (state ``Prd type''), \rev{the next token can either be} an $\mathrm{x}_1$ coordinate for another product, or a $\texttt{[Prd]}$ token if all the products have been predicted. Such constraints are enforced by masking the output vocabulary to avoid generating an invalid token, \rev{i.e., the model can only generate tokens that follow the sequence representation grammar, and the invalid tokens are masked out.}
Finally, the predicted sequence is converted to the reaction structure according to its definition. 

\subsection{Data}

To train and evaluate RxnScribe, we create a high-quality dataset with annotated reaction structures over diagrams extracted from chemistry literature. \Cref{tab:data} summarizes the statistics of our dataset.

\begin{table}[t]
    \centering
    \begin{tabular}{l|cccc|c}
        \toprule
            & Single-line   & Multiple-line & Tree  & Graph & Overall \\\midrule
        Num.~of diagrams    
            & 730           & 260           & 286   & 102   & 1378 \\
        Num.~of entities
            & 7536          & 4831          & 4934  & 1926  & 19227 \\
        Num.~of reactions 
            & 882           & 948           & 1313  & 633   & 3776  \\
        Avg.~num.~of reactions per diagram
            & 1.2           & 3.6           & 4.6   & 6.2   & 2.7 \\
        \bottomrule
    \end{tabular}
    \caption{Statistics of our reaction diagram parsing dataset.}
    \label{tab:data}
\end{table}

\begin{figure}[t]
    \centering
    \includegraphics[width=0.6\linewidth]{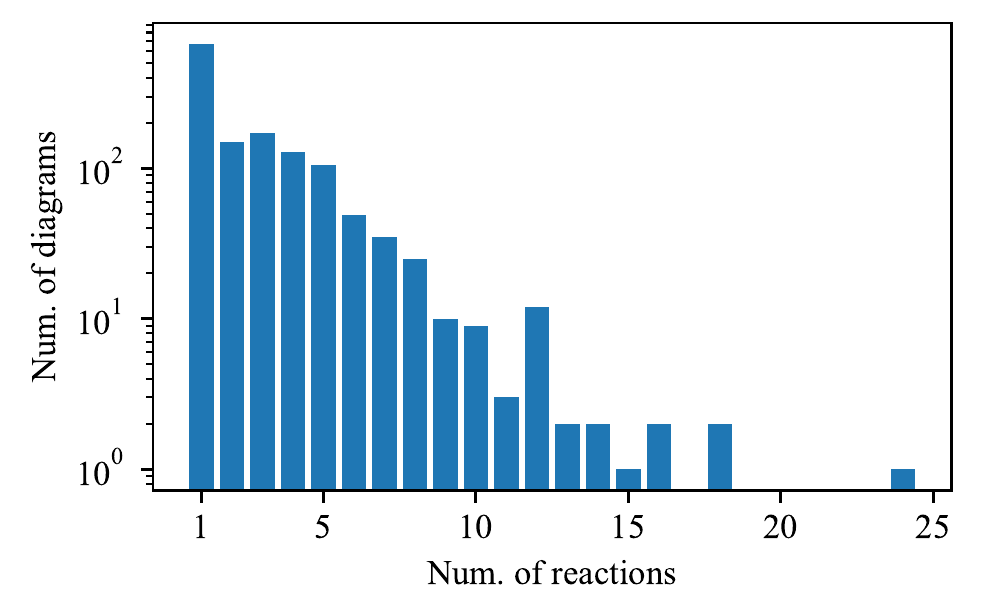}
    \caption{The number of diagrams with respect to the number of reactions in each diagram.}
    \label{fig:num_reaction}
\end{figure}

\paragraph{Dataset Construction}
We collect a list of 662 articles from four chemistry journals: Journal of the American Chemical Society, Journal of Organic Chemistry, Organic Letters, and Organic Process Research \& Development, where each article is a PDF file. We use the pdffigures tool\cite{pdffigures} to extract the diagram images from the PDF files. The diagrams are categorized into four styles: single-line, multiple-line, tree, and graph, based on how the reactions are organized in the diagram. An example of each style can be found in \Cref{fig:diagram}. In \Cref{fig:num_reaction}, we show the distribution of the number of reactions in each diagram. The majority of the diagrams contain fewer than 10 reactions, and about half of all examples are simple diagrams with a single reaction.

\paragraph{Annotation}
The annotation process consists of two stages: entity annotation and reaction role annotation. An example of the annotation can be found in the supporting information.

The first step is to annotate the relevant entities in the diagrams, typically presented in the form of a chemical structure (molecular graph) or a text sequence (chemical name and formula, identifier, etc).
For each entity, we annotate a rectangle bounding box (defined by four coordinates) and its entity type. As mentioned in the model section, we consider three main types of entities: molecule, text, and structure identifier. Amazon Mechanical Turk (MTurk) was used to collect the initial entity annotations. 
We carefully refined them using the CVAT \cite{cvat} platform to resolve annotation errors and ambiguities, which mainly involve bounding box tightness and missing entities. For example, small-sized entities such as structure identifiers were often skipped by the annotators. Finally, 23\% (317) diagrams have been manually corrected.
In this work, we only annotate the bounding boxes of the entities, and do not annotate the SMILES strings of the molecules or the content of the texts due to the high annotation costs.

The second step is to annotate the reaction roles given the diagram and the annotated entities. As this process requires domain knowledge, two students with bachelor's degrees in chemistry performed the annotation. Each annotator was given the diagram with visualized entities, and each entity was associated with a unique index. All possible reactions were annotated in a sequential form, and each reaction was annotated with the three reaction roles (\texttt{[Rct]}, \texttt{[Cnd]}, \texttt{[Prd]}). The annotation followed three general guidelines:
\begin{enumerate}
    \item All reactions displayed in the diagram, including intermediate steps, should be annotated. However, if the diagram indicates a reaction is not valid (e.g., a cross mark on the arrow, occurs less than 5\% in the dataset), it is not annotated. 
    \item For each reaction, all of its reactant/condition/product entities should be annotated as specified in the diagram. Conditions include reagent, catalyst, solvent, temperature, and time. We also annotate other relevant information such as reaction type as conditions. \rev{In cases where byproducts are present, we do not distinguish them with the main product and annotate both as products.}
    \item Annotation follows the reading order in general, i.e., top-to-bottom and left-to-right; however, for tree and graph-style diagrams where there is not a natural reading order, we do not specify a particular order and leave the decision to the annotator.
\end{enumerate}
Finally, one author of the paper double-checked the annotations to guarantee their correctness and consistency.

\subsection{Data Augmentation}
Given the constructed dataset, we are able to train a neural network model for reaction diagram parsing. However, the number of annotated diagrams is relatively small, and about half of them are simple and consist of a single reaction (see \Cref{fig:num_reaction}). Because our goal is to train a robust model that accurately parses diagrams with different styles in the real world, we develop data augmentation techniques to generate synthetic reaction diagrams during training.

\begin{figure}[t]
    \centering
    \includegraphics[width=\linewidth]{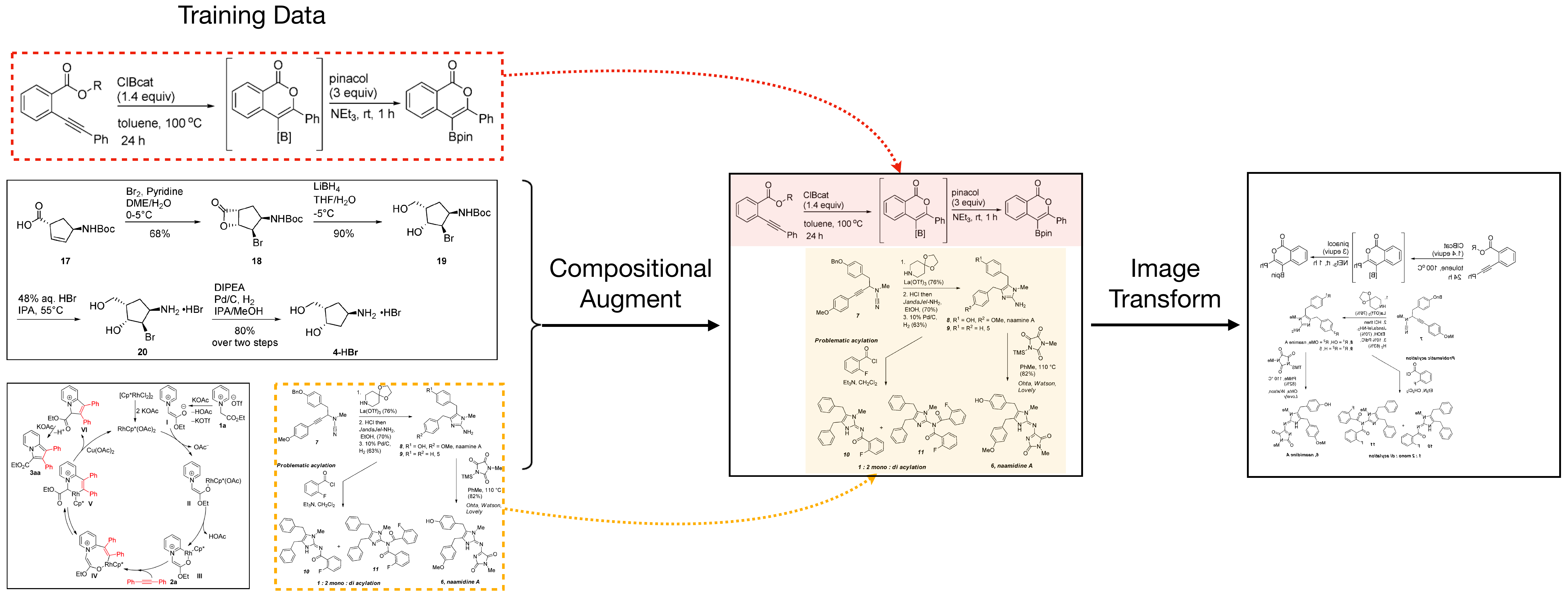}
    \caption{Data augmentation in \ours. We randomly compose simple diagrams into more complex ones, and apply image transformations.}
    \label{fig:augment}
\end{figure}

\Cref{fig:augment} illustrates the data augmentation process. We first have a \textit{compositional augmentation} stage to synthesize more complex diagrams from the training data. 
Specifically, we randomly sample multiple diagrams from the training data, and concatenate them vertically to form a new diagram with multiple reactions. If the diagrams have different width, a random offset is added to the shorter one. The number of diagrams that are concatenated together ranges from 2 to 6, with the probability of concatenating more diagrams decreasing exponentially. The annotation of the new diagram is the combination of the annotations of the original diagrams, while the entity bounding boxes are shifted accordingly in the new diagram. 
After compositional augmentation, we further apply image transformations, including resizing, padding, rotation, flipping, and color jitter at random. These transformations improve the model's robustness against image perturbations. 

\section{Experiments}

\subsection{Experimental Setup}
\rev{ The model architechture of \ours consists of a ResNet-50 backbone and a 6-layer Transformer for sequence generation. This architecture was  adoped in previous research on object detection\cite{CarionMSUKZ20,pix2seq}. We initialize the parameters of \ours with a Pix2Seq model checkpoint\cite{pix2seq} pre-trained on the MS-COCO object detection dataset\cite{mscoco}, which contains 118K images and the annotations of object bounding boxes. We finetune the model on our dataset for 600 epochs,} with a maximum learning rate of 3e-4, a linear warmup for the first 2\% steps and a cosine function decay. The training batch size is 32. The input diagram is resized and padded to a fixed resolution of 1333$\times$1333. At inference time, we post-process the predictions to remove obvious mistakes, such as duplicate reactions and empty entities. 

Due to the small size of the dataset, we perform 5-fold cross validation to evaluate \ours. The dataset is evenly split into five subsets. We train five models, each model uses four subsets for training and development, and the other one for testing. We report the  evaluation results on the combined five test sets.

We compare \ours with two reaction diagram parsing software, ReactionDataExtractor\cite{rde} and OChemR\cite{OChemR}. ReactionDataExtractor is a rule-based pipeline. OChemR trains an object detection model to recognize arrows, molecules, and texts, and uses heuristics to identify the reaction roles.

\subsection{Evaluation Metric}
\begin{figure}[t!]
    \centering
    \includegraphics[width=\linewidth]{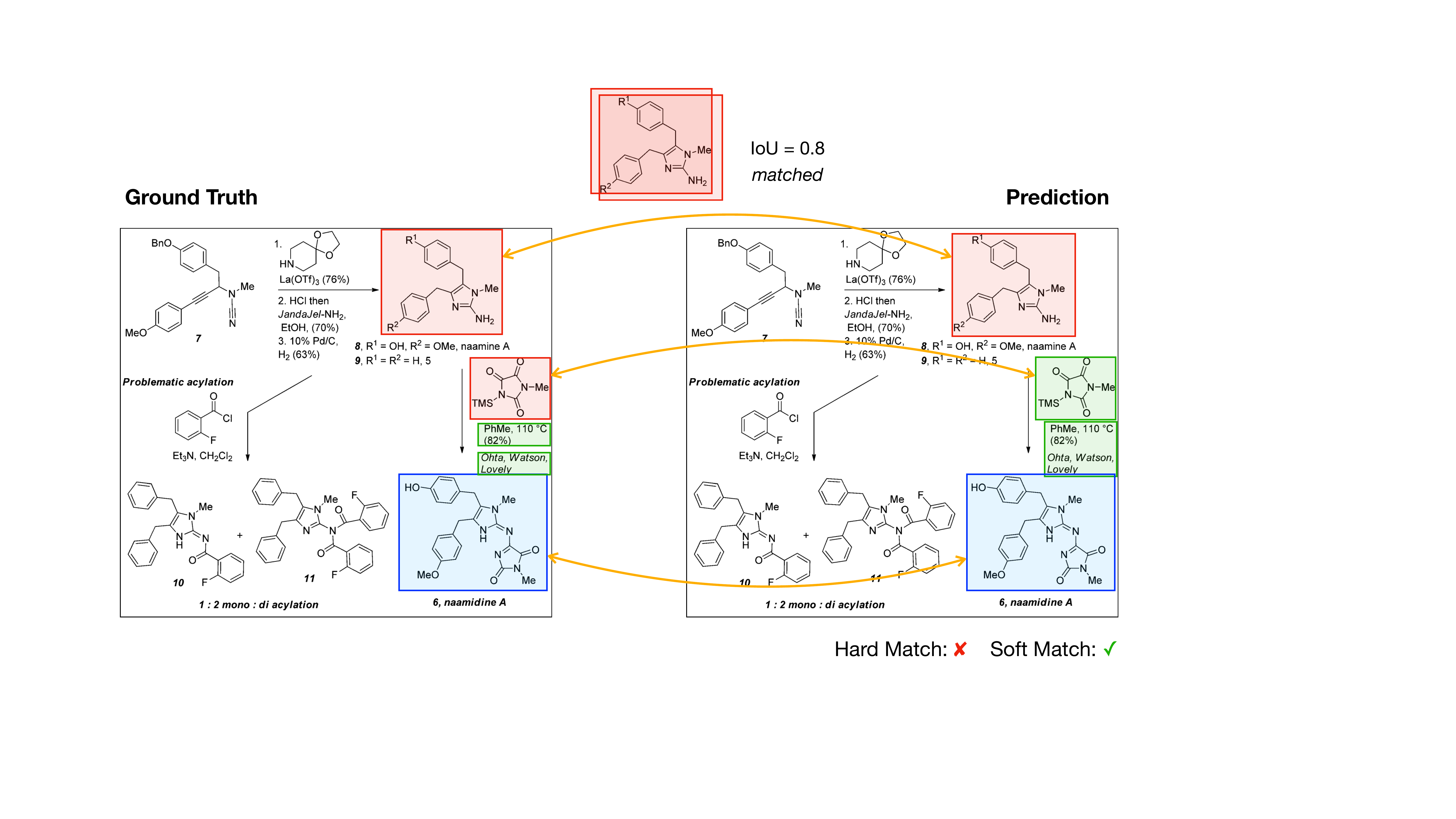}
    \caption{Evaluation of reaction diagram parsing results. Each ground truth entity  is matched with a predicted entity that has the maximum IoU overlap (at least 0.5). Hard match evaluation requires all the reactants, conditions, and products to be matched. Soft match evaluation only considers molecule entities and does not distinguish between reactants and reagents (part of the conditions).}
    \label{fig:eval}
\end{figure}

The evaluation of reaction diagram parsing results is non-trivial, as both prediction and ground truth are sets of reaction structures. Often, the prediction does not match the ground truth exactly. For example, the entity bounding boxes may be slightly shifted or the orders of the reactions predicted differently, but many of these cases should be considered as correct. We design two groups of evaluation metrics, hard match and soft match, to evaluate the model. \Cref{fig:eval} illustrates the evaluation process, which will be explained in detail next.

For each diagram, we denote the ground truth as $\mathcal{G}=\{R_1, R_2, \dots, R_n\}$ and the prediction as $\mathcal{P}=\{\hat{R}_1, \hat{R}_2, \dots, \hat{R}_m\}$. We first describe how to compare a predicted reaction $\hat{R}$ with a ground truth reaction $R$. We find a mapping between the two lists of entities in $\hat{R}$ and $R$. Specifically, for each entity in $R$, we find the entity in $\hat{R}$ which has the maximum bounding box overlap with it. The bounding box overlap is measured by the Intersection over Union (IoU) score. If the maximum IoU is greater than a threshold $0.5$, we consider the predicted and ground truth bounding boxes as being successfully matched. 

In our \textit{hard match} evaluation, we say the prediction $\hat{R}$ matches the ground truth $R$ if all the reactants, conditions, and products of the $\hat{R}$ and $R$ can be matched. In the \textit{soft match} evaluation, we only take into account the molecule entities, and do not distinguish between reactants and reagents (part of the conditions). We have two considerations for the soft match evaluation. First, it only compares molecule entities and ignores text entities, because there is often ambiguity in whether two consecutive text lines are annotated as one entity or two entities. Second, it does not distinguish reactants and conditions, because sometimes a molecule is drawn above or below the reaction arrow and visually looks like a condition, but actually contributes heavy atoms and is conventionally considered a reactant. \rev{\Cref{fig:eval} provides an illustrative example, where the molecule containing a TMS group is annotated as a reactant in the ground truth, but predicted as a reaction condition by the model. While this prediction is considered incorrect under hard match evaluation, it is considered correct under soft match evaluation. } 

For both hard match and soft match, we compute the precision, recall, and F1 scores. As we do not have the one-to-one correspondence between the predicted reactions and the ground truth reactions, we enumerate all pairs and compare each $\hat{R}_i$ with each $R_j$. Then the metrics are defined as follows:
\begin{equation}
\begin{split}
    \mathrm{precision} &= \frac{1}{m} \sum_{j=1}^m \mathbbm{1}\left(\exists~i\in \{1,\dots,n\}, \hat{R}_j \mathrm{~matches~} R_i\right) \\
    \mathrm{recall} &= \frac{1}{n} \sum_{i=1}^n \mathbbm{1}\left(\exists~j\in \{1,\dots,m\}, R_i \mathrm{~matches~} \hat{R}_j\right) \\
    \textrm{F1} &= \frac{2\cdot \mathrm{precision}\cdot\mathrm{recall}}{\mathrm{precision} + \mathrm{recall}}.
\end{split}
\end{equation}
The precision measures what fraction of the model predictions are correct, and the recall measures what fraction of the ground truth are correctly predicted. Finally, we report the micro-averaged metrics over the test set.

\subsection{Results and Analysis}

\begin{table}[t!]
    \centering
    \begin{tabular}{lccccccc}
    \toprule
        & \multicolumn{3}{c}{Hard Match} && \multicolumn{3}{c}{Soft Match} \\ \cmidrule{2-4} \cmidrule{6-8}
        & Precision & Recall & F1 && Precision & Recall & F1 \\\midrule
    ReactionDataExtractor & 4.1 & 1.3 & 1.9 & & 19.4 & 5.9 & 9.0 \\
    OChemR & 4.4 & 2.8 & 3.4 & & 12.4 & 7.9 & 9.6 \\\midrule
    RxnScribe & \textbf{72.3} & \textbf{66.2} & \textbf{69.1} & & 83.8 & \textbf{76.5} & \textbf{80.0} \\
    \quad - No pre-training & 66.4 & 59.4 & 62.7 & & 80.4 & 71.3 & 75.5 \\
    \quad - No compositional augmentation & 67.1 & 60.7 & 63.8 & & 78.2 & 70.2 & 74.0 \\
    \quad - Random reaction order & 72.0 & 64.2 & 67.9 & & \textbf{83.9} & 74.3 & 78.8 \\
    \quad - No post-processing & 70.8 & 66.0 & 68.3 & & 82.1 & 76.4 & 79.1 \\
    \bottomrule
    \end{tabular}
    \caption{Evaluation of reaction diagram parsing performance (scores are in \%).}
    \label{tab:overall_res}
\end{table}

\Cref{tab:overall_res} shows the overall evaluation results. As the first neural model for reaction diagram parsing, \ours achieves strong performance (hard match F1 69.1\% and soft match F1 80.0\%). We present four ablation studies in \Cref{tab:overall_res}: (1) training the model from scratch without pre-training on object detection; (2) training without compositional augmentation; (3) using a random order of reactions for each diagram instead of the reading order (i.e., the annotation order); (4) without post-processing at inference time. All these variants perform worse than the full model. \ours leverages the proposed techniques to achieve strong performance with limited training data. We notice that \ours sometimes predicts duplicate reactions or empty entities, and these mistakes have been removed with a simple post-processing which leads to about 1\% improvement in F1.

\begin{figure}[t!]
    \centering
    \includegraphics[width=\linewidth]{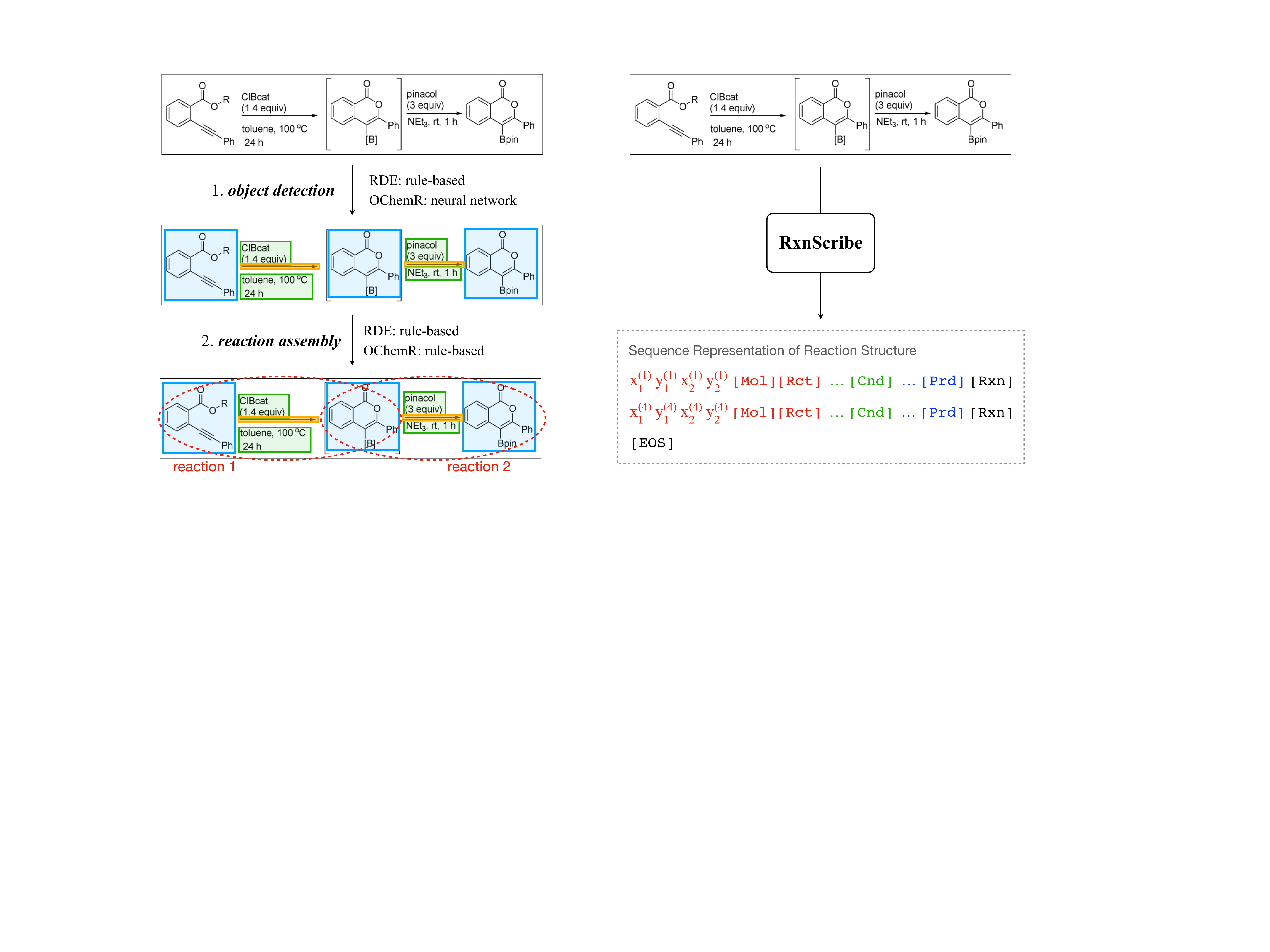}
    \caption{\rev{Comparison between previous pipelined approaches (left) and \ours (right). RDE: ReactionDataExtractor.}}
    \label{fig:compare}
\end{figure}

\begin{figure}[t!]
    \centering
    \includegraphics[width=\linewidth]{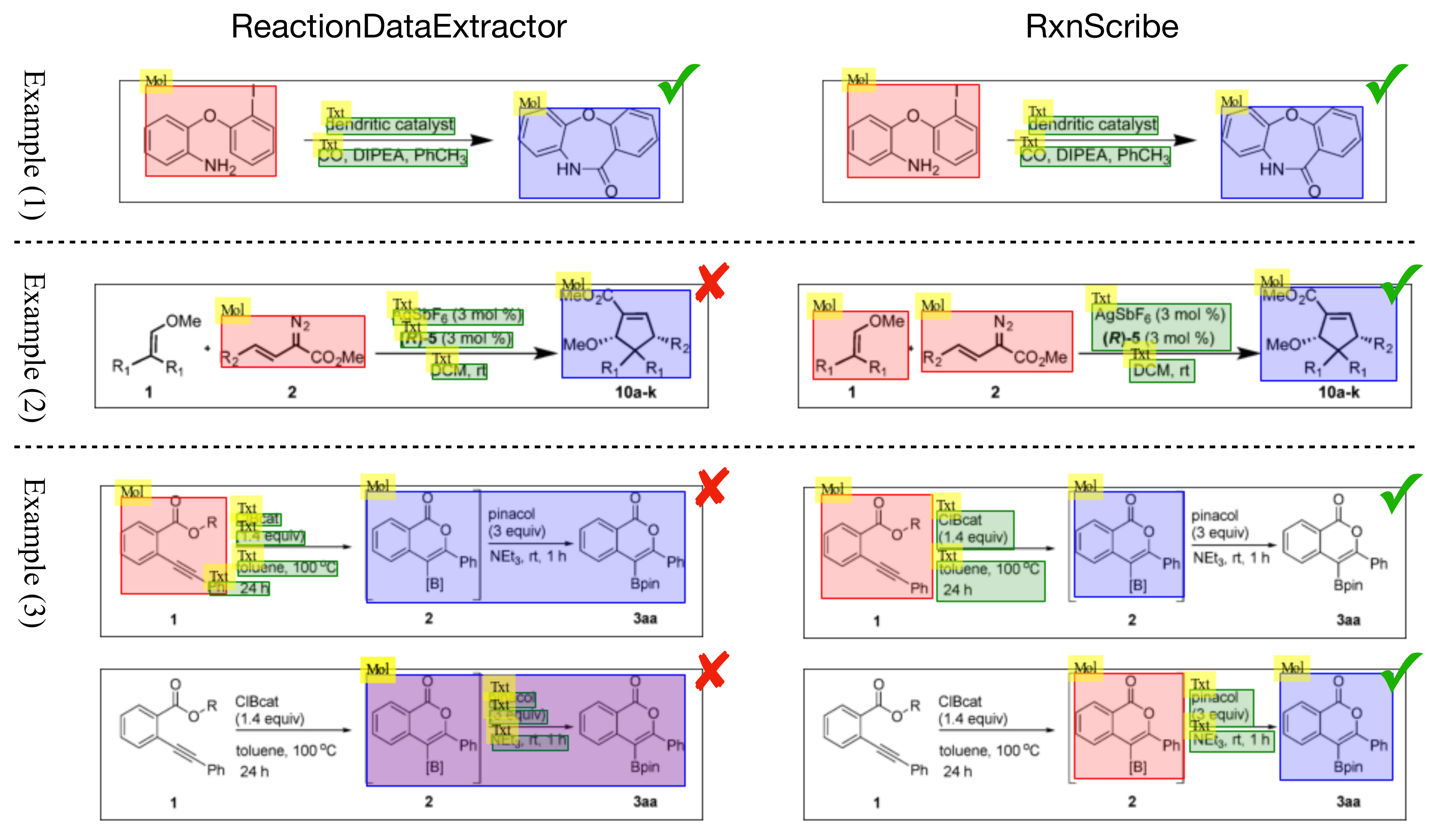}
    \caption{Examples of ReactionDataExtractor (left) and \ours (right) predictions. We analyze three single-line diagrams in our dataset, which are adapted with permission from 
    \citet{lu2005intramolecular} [Copyright © 2005 American Chemical Society],
    \citet{briones2013enantioselective} [Copyright © 2013 American Chemical Society],
    and
    \citet{faizi2016catalyst} [Copyright © 2016 American Chemical Society]. 
    Check marks and cross marks represent correct and wrong predictions, respectively.}
    \label{fig:rde_pred}
\end{figure}

\ours markedly outperforms existing rule-based systems, whose soft match F1 scores are below 10\%, at least in part because they have not been tuned for the diagrams in our dataset. For example, ReactionDataExtractor designed its rules based on a dataset that contains mostly single-line diagrams. In our evaluation, ReactionDataExtractor achieves 27.4\% precision, 15.0\% recall, and 19.4\% F1 (soft match) on single-line diagrams. \Cref{fig:rde_pred} compares ReactionDataExtractor and \ours's predictions on three single-line diagrams. In the first example, where there is a single reactant and a single product, and the separation between the molecules and the arrow is clear,  ReactionDataExtractor can recognize it correctly. In the other two examples, where the reaction involves multiple reactants or the diagram contains multiple reactions, ReactionDataExtractor makes mistakes such as missing a reactant or incorrectly merging two molecule entities. Our \ours model handles these cases adequately, and achieves a 91.0\% soft match F1 on single-line diagrams.

\ours's success can be credited to its sequence generation formulation, which avoids the inherent problems of the previous pipelined approach,
i.e., first segment out the relevant entities and then compose them into reactions. \rev{\Cref{fig:compare} compares the two solutions.}
\rev{Our method has three advantages compared with the pipelined approach. First, our method avoids the issue of error propagation, while in a pipelined approach, inaccurate entity detection can lead to compounded errors in the subsequent steps. Second, our neural network model can generalize well to diverse reaction diagrams, without relying on complex heuristics to assemble entities into reactions. Third, we do not require the annotation of the arrows or the association between arrows and the entities. \ours directly generates the reaction structure as a sequence, skipping the intermediate step of arrow and entity detection. }

\begin{figure}[p!]
    \centering
    \vspace{-0.16in}
    \includegraphics[width=0.92\linewidth]{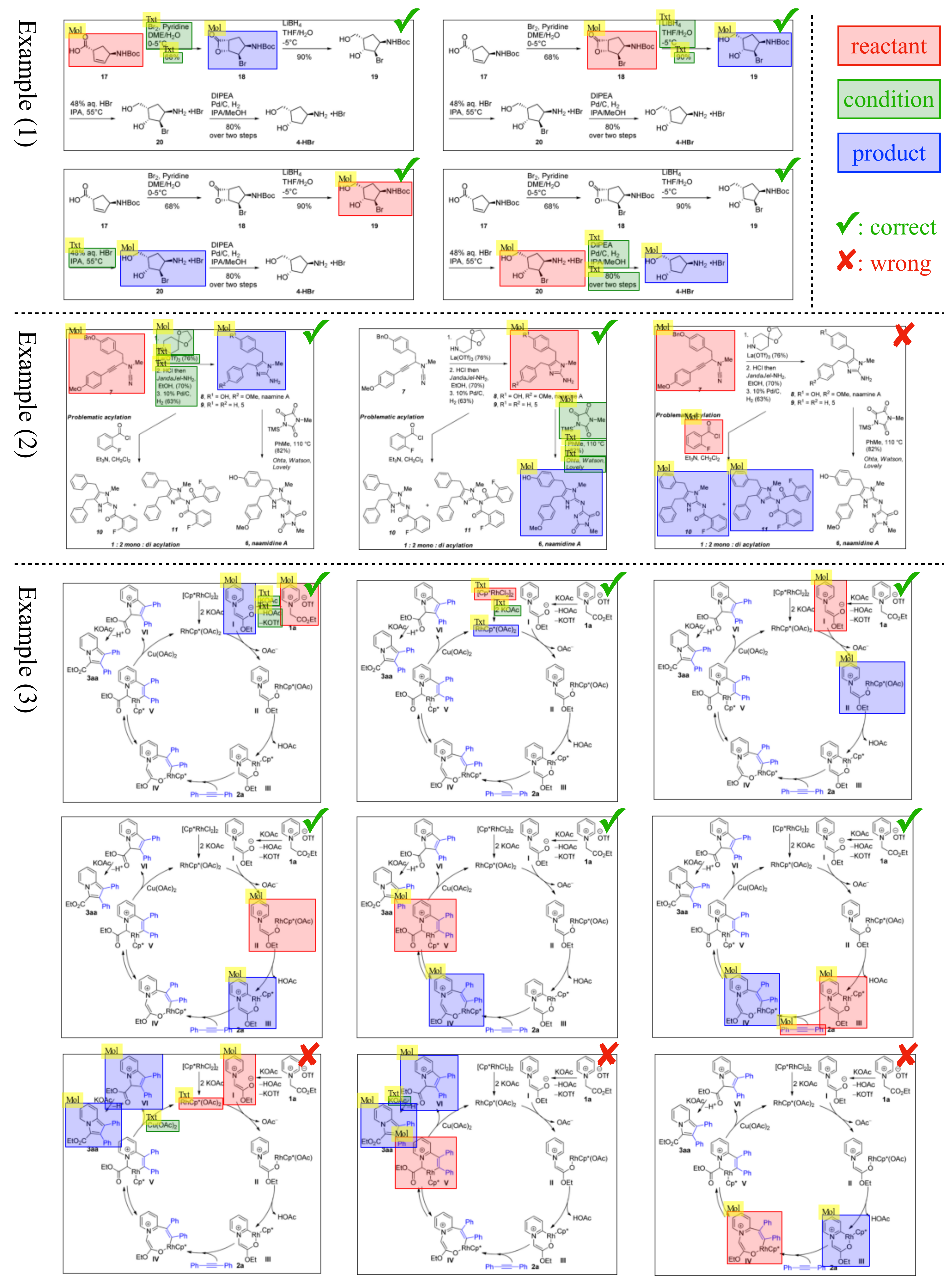}
    \caption{Examples of \ours predictions. Each predicted reaction is visualized in a separate image. The original diagrams are shown in \Cref{fig:diagram}. Check marks and cross marks represent correct and wrong prediction, respectively, under the soft match criterion.}
    \label{fig:pred_example}
\end{figure}

\Cref{fig:pred_example} shows some examples of \ours's prediction on more complex diagrams. Example (1) is a multiple-line diagram, where \ours correctly predicts all four reactions. Example (2) is a tree-style diagram with three reactions. \ours predicts one horizontal and one vertical reaction correctly, but makes a mistake on the third reaction, probably because the arrow has two lines which is rare in the dataset. Example (3) is a graph-style diagram with nine annotated reactions. \ours makes six correct predictions and three wrong predictions, and there are three other reactions in the ground truth not predicted by the model. This diagram contains many curved and branched arrows, which are still challenging for \ours.

\begin{figure}[t]
    \centering
    \includegraphics[width=\linewidth]{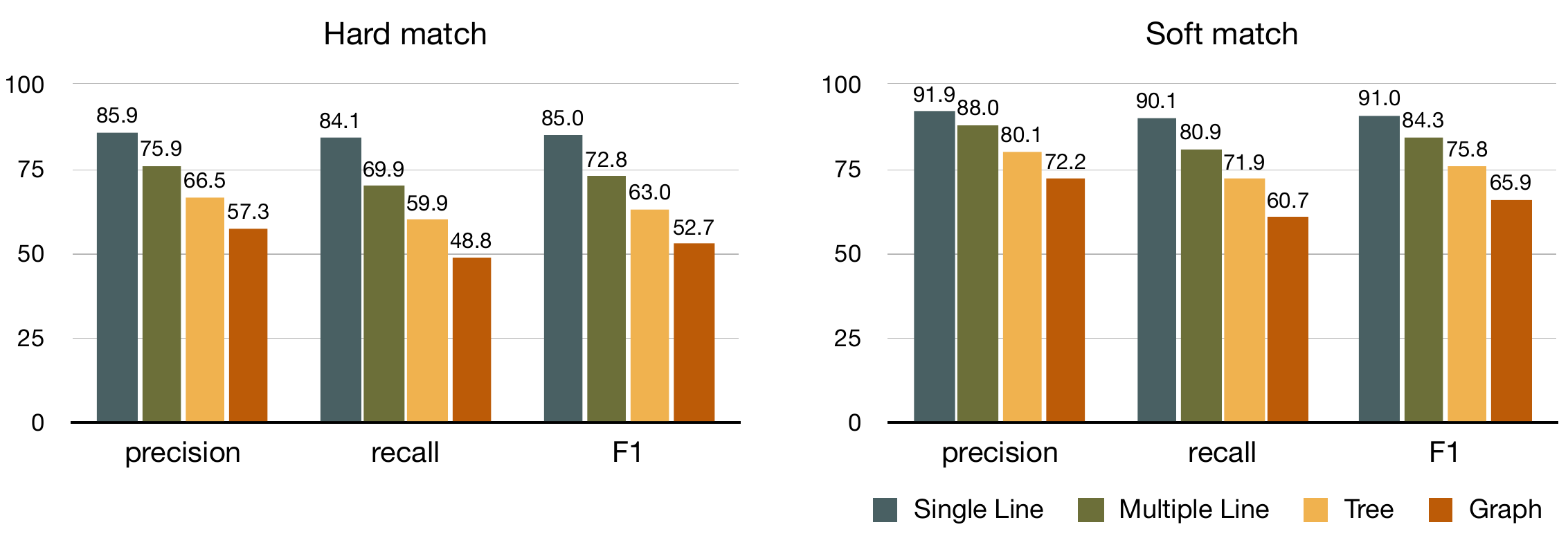}
    \caption{\ours's performance on four styles of diagrams.}
    \label{fig:diagram_res}
\end{figure}

\begin{figure}[t!]
    \centering
    \includegraphics[width=\linewidth]{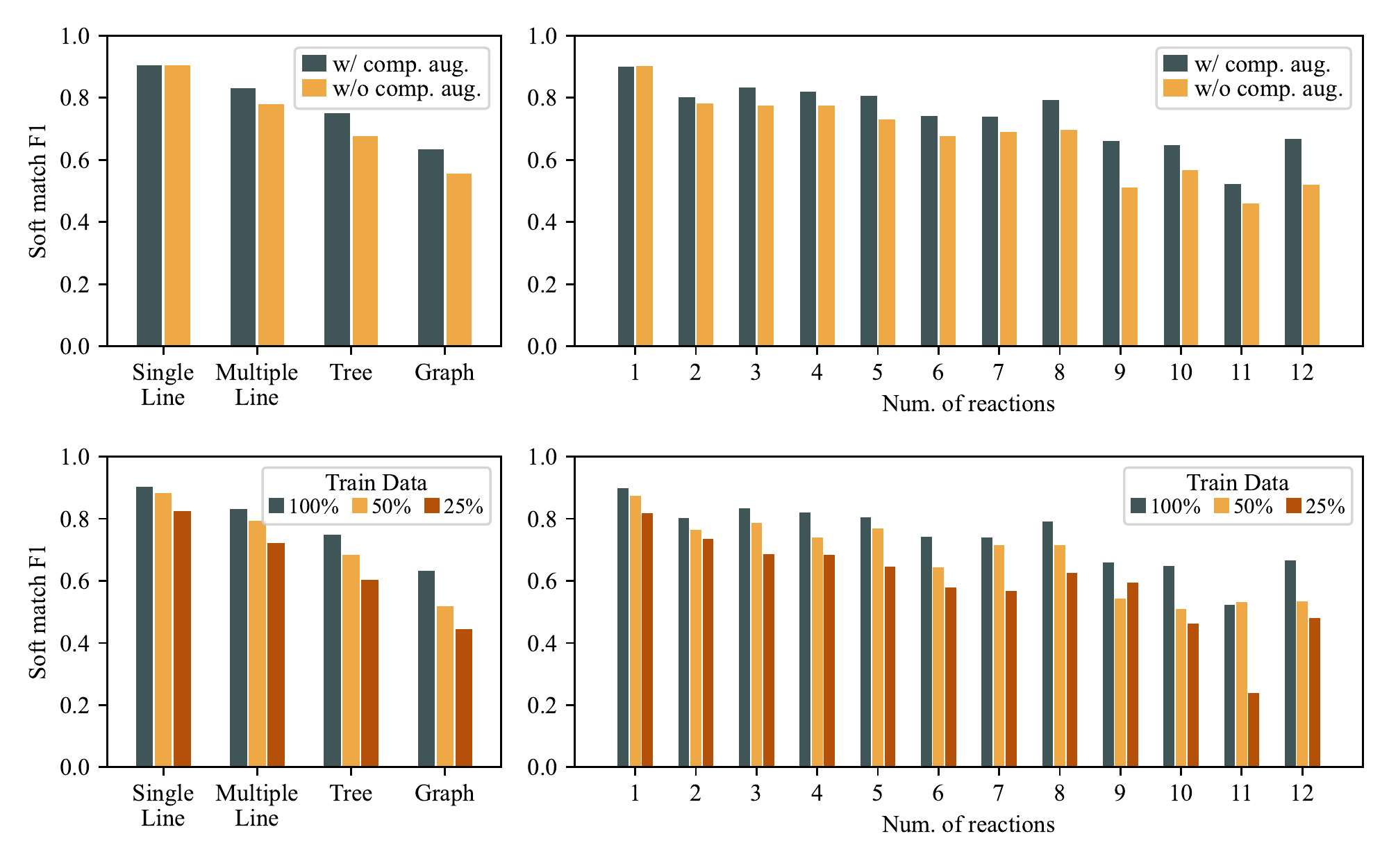}
    \caption{Top: performance of models trained with or without compositional augmentation. Bottom: performance of models trained with 100\%, 50\%, or 25\% of training data.}
    \label{fig:num_rxn_res}
\end{figure}

\Cref{fig:diagram_res} decomposes \ours's performance on four diagram styles. The model performs the best on single-line diagrams, achieving a 90.4\% soft match F1 score, but performs relatively worse on the other three styles. It is expected because the other three styles are more complex, diverse, and contain more reactions in average. Besides, there are fewer such examples than single line diagrams in the training data, which leads to lower accuracy on those diagrams. Nevertheless, even in the hardest group ``graph'', which only has 102 examples in the dataset, \ours still achieves above 60\% soft match F1.

In \Cref{fig:num_rxn_res}, we show how the compositional augmentation and using more training data help the model to achieve better performance. We analyze the results on the four styles, as well as the diagrams with different number of reactions. On single line diagrams and those with only one reaction, compositional augmentation does not help because it only creates diagrams with multiple lines and multiple reactions, but the performance on the other three styles have been significantly improved with this augmentation. Comparing the models trained with 25\%, 50\%, and 100\% data, we observe the performance constantly improves with more training data, especially on complex diagrams with multiple reactions. It implies the potential of further boosting the performance by collecting more training data. \rev{A promising avenue for future research is to adopt an active learning strategy, whereby we use the predictions of our current \ours model to selectively identify and annotate more challenging diagrams.}

\section{Conclusion}
This paper presents RxnScribe, a novel model for reaction diagram parsing. We define a sequence representation to describe the reaction structure in a diagram, where each entity, reaction role, and reaction is expressed as a sequence of discrete tokens. \ours leverages this simple and effective formulation and trains a sequence generation model to predict the reaction structure. 
We collect a dataset of 1,378 diagrams to train and evaluate \ours. Our experiments validate that \ours can accurately parse the reaction diagrams in different styles. \rev{Our model's performance on specific types of reactions can be further improved by annotating more diagrams, such as biosynthesis and metabolic pathways, which will facilitate data extraction in these domains.}

We contribute to this research area by defining the task of reaction diagram parsing, proposing the first machine learning solution, and constructing a diverse dataset for training and evaluation. Despite the success in our experiments, there are a few limitations in this work. First, we focus on parsing the reaction structure, but do not evaluate the final extracted reaction SMILES strings due to the lack of such ground truth in our dataset. The molecular structure recognition model MolScribe and the OCR tool might introduce additional errors. Second, we limit our study to diagrams presented in digital format, excluding those that are either scanned or hand-drawn. Furthermore, the extracted information from diagrams are sometimes incomplete. For example, the reaction conditions are sometimes listed in a table and the molecules may involve R-groups which are defined elsewhere. Future work needs to design methods to consolidate the information from diagrams, tables, and texts.

\section{Data and Software Availability}
Our code, data, and model checkpoints are publicly available at \url{https://github.com/thomas0809/RxnScribe}. We have also developed a web interface for \ours: \url{https://huggingface.co/spaces/yujieq/RxnScribe}. Our dataset is constructed on the journal articles shared between the American Chemical Society (ACS) and MIT under a private access agreement. We have obtained approval from ACS to release the dataset for future research.

\begin{suppinfo}

Detailed evaluation results for \ours and other tools, an illustration of our data annotation process, and the sources of our reaction diagram dataset are available in the supporting information.

\end{suppinfo}

\begin{acknowledgement}
The author thanks the members of Regina Barzilay's Group at MIT CSAIL for helpful discussion and feedback. This work was supported by the DARPA Accelerated Molecular Discovery (AMD) program under contract HR00111920025 and the Machine Learning for Pharmaceutical Discovery and Synthesis Consortium (MLPDS).

\end{acknowledgement}

\bibliography{reference}


\end{document}


\section{Experiment Results}

\Cref{tab:hard} and \Cref{tab:soft} display the complete evaluation results of \ours and the compared models. \Cref{tab:hard} shows the hard match scores and \Cref{tab:soft} shows the soft match scores. We present both the overall performance and the performance on each diagram style. \ours achieves better performance compared to other models. 

\begin{sidewaystable}[ph!]
    \centering
    \small
    \setlength{\tabcolsep}{4pt}
    \caption{Hard match evaluation results (scores are in \%).}
    \label{tab:hard}
    \begin{tabular}{lccccccccccccccccccc}
    \toprule
        & \multicolumn{3}{c}{Overall} && \multicolumn{3}{c}{Single-Line} && \multicolumn{3}{c}{Multiple-Line} && \multicolumn{3}{c}{Tree} && \multicolumn{3}{c}{Graph} \\ \cmidrule{2-4} \cmidrule{6-8} \cmidrule{10-12} \cmidrule{14-16} \cmidrule{18-20}
        & Prec. & Recall & F1 && Prec. & Recall & F1 && Prec. & Recall & F1 && Prec. & Recall & F1 && Prec. & Recall & F1 \\\midrule
    ReactionDataExtractor       & 4.1  & 1.3  & 1.9  && 6.2  & 3.4  & 4.4  && 2.8  & 0.9  & 1.4 &
                                & 2.8  & 0.5  & 0.9  && 2.0  & 0.3  & 0.5 \\
    OChemR                      & 4.4  & 2.8  & 3.4  && 2.6  & 2.7  & 2.6  && 5.1  & 3.6  & 4.2 &
                                & 4.7  & 2.1  & 2.9  && 9.4  & 3.2  & 4.7
    \\\midrule
    RxnScribe                   & 72.3 & 66.2 & 69.1 && 85.9 & 84.1 & 85.0 && 75.9 & 69.9 & 72.8 &
                                & 66.5 & 59.9 & 63.0 && 57.3 & 48.8 & 52.7 \\
    \quad - Not pre-trained     & 66.4 & 59.4 & 62.7 && 82.1 & 81.7 & 81.9 && 73.4 & 66.9 & 70.0 & 
                                & 58.1 & 51.9 & 54.8 && 44.4 & 32.9 & 37.8 \\
    \quad - No compos. aug.     & 67.1 & 60.7 & 63.8 && 83.3 & 82.7 & 83.0 && 69.5 & 63.9 & 66.6 & 
                                & 59.8 & 53.2 & 56.3 && 51.9 & 40.9 & 45.8 \\
    \quad - Random order        & 72.0 & 64.2 & 67.9 && 84.9 & 83.6 & 84.2 && 75.9 & 67.9 & 71.7 &
                                & 65.7 & 58.1 & 61.7 && 57.1 & 44.4 & 50.0 \\
    \bottomrule
    \end{tabular}
    \vspace{0.8in}
    \caption{Soft match evaluation results (scores are in \%).}
    \label{tab:soft}
    \begin{tabular}{lccccccccccccccccccc}
    \toprule
        & \multicolumn{3}{c}{Overall} && \multicolumn{3}{c}{Single-Line} && \multicolumn{3}{c}{Multiple-Line} && \multicolumn{3}{c}{Tree} && \multicolumn{3}{c}{Graph} \\ \cmidrule{2-4} \cmidrule{6-8} \cmidrule{10-12} \cmidrule{14-16} \cmidrule{18-20}
        & Prec. & Recall & F1 && Prec. & Recall & F1 && Prec. & Recall & F1 && Prec. & Recall & F1 && Prec. & Recall & F1 \\\midrule
    ReactionDataExtractor       & 19.4 & 5.9  & 9.0  && 27.4 & 15.0 & 19.4 && 11.0 & 3.6 & 5.4 &
                                & 19.8 & 3.7  & 6.3  && 6.9  & 1.1  & 1.9 \\
    OChemR                      & 12.6 & 8.0  & 9.8  && 9.8  & 10.5 & 10.1 && 12.5 & 8.8 & 10.3 &
                                & 15.3 & 6.7  & 9.3  && 17.9 & 6.0  & 9.0 
    \\\midrule
    RxnScribe                   & 83.8 & 76.5 & 80.0 && 91.9 & 90.1 & 91.0 && 88.0 & 80.9 & 84.3 &
                                & 80.1 & 71.9 & 75.8 && 72.2 & 60.7 & 65.9 \\
    \quad - Not pre-trained     & 80.4 & 71.3 & 75.5 && 90.9 & 90.5 & 90.7 && 86.7 & 78.9 & 82.6 &
                                & 73.5 & 64.6 & 68.7 && 62.2 & 46.9 & 54.9 \\
    \quad - No compos. aug.     & 78.2 & 70.2 & 74.0 && 90.9 & 90.1 & 90.5 && 81.4 & 74.8 & 78.0 &
                                & 72.3 & 63.7 & 67.7 && 64.1 & 49.0 & 55.5 \\
    \quad - Random order        & 83.9 & 74.3 & 78.8 && 91.9 & 90.5 & 91.2 && 87.0 & 77.8 & 82.2 &
                                & 79.9 & 69.7 & 74.5 && 73.6 & 55.9 & 63.5 \\
    \bottomrule
    \end{tabular}
\end{sidewaystable}

\section{Annotation}

\begin{figure}[t!]
    \centering
    \includegraphics[width=0.99\linewidth]{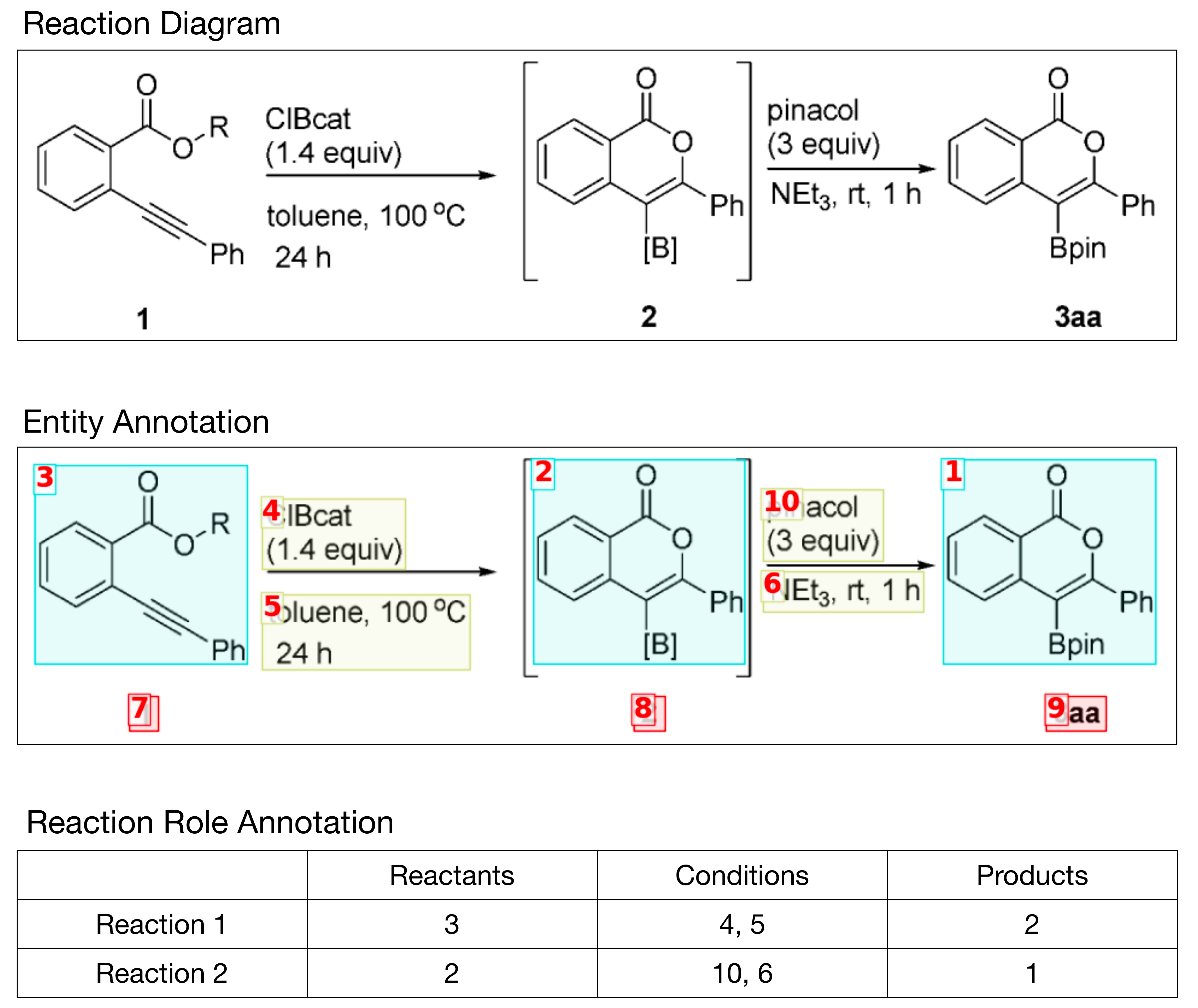}
    \caption{Our annotation process. First, we annotate the entities in a diagram, and assign an index to each entity. Second, we annotate the reaction roles. The example diagram is from a journal article.\cite{faizi2016catalyst}}
    \label{fig:annot}
\end{figure}

\Cref{fig:annot} shows an example of our two-step annotation procedure. 

Given a reaction diagram, we first annotate the entities, including molecules, texts, and identifiers, and assign indices to them. Each entity is annotated with its bounding box and entity type. The entity annotation was performed by was performed by Amazon Mechanical Turk and cost about \$1000.

Then, we visualize the entity bounding boxes and annotate the reaction roles. For each reaction, the annotator selects the indices of its reactants, conditions, and products. This annotation process was performed by two chemistry students and took approximately two months to complete. 

\newpage 

\section{Dataset}
Our dataset contains 1,378 reaction diagrams, collected from 662 journal articles shared by ACS. We have obtained approval from ACS to release the dataset. The diagrams can be downloaded at \url{https://huggingface.co/yujieq/RxnScribe/blob/main/images.zip}. The ground truth files are included in our GitHub repository (\url{https://github.com/thomas0809/RxnScribe/tree/main/data/parse}).

We list the DOI numbers of the relevant journal articles below.

{
\small
\setlength{\tabcolsep}{12pt}
\begin{longtable}{lllll}
acs.joc.5b00301 & acs.joc.5b00302 & acs.joc.5b00632 & acs.joc.5b00685 \\
acs.joc.5b01204 & acs.joc.5b01366 & acs.joc.5b01547 & acs.joc.5b01703 \\
acs.joc.5b02057 & acs.joc.5b02237 & acs.joc.5b02345 & acs.joc.5b02382 \\
acs.joc.6b00020 & acs.joc.6b00116 & acs.joc.6b01001 & acs.joc.6b01262 \\
acs.oprd.5b00027 & acs.oprd.5b00070 & acs.oprd.5b00137 & acs.oprd.5b00144 \\
acs.oprd.5b00148 & acs.oprd.5b00170 & acs.oprd.5b00209 & acs.oprd.5b00251 \\
acs.oprd.5b00278 & acs.oprd.5b00282 & acs.oprd.5b00303 & acs.oprd.5b00312 \\
acs.oprd.5b00331 & acs.oprd.5b00339 & acs.oprd.5b00370 & acs.oprd.5b00371 \\
acs.oprd.5b00379 & acs.oprd.5b00418 & acs.oprd.6b00011 & acs.oprd.6b00095 \\
acs.oprd.6b00117 & acs.oprd.6b00126 & acs.oprd.6b00128 & acs.oprd.6b00180 \\
acs.oprd.6b00188 & acs.orglett.5b00081 & acs.orglett.5b00312 & acs.orglett.5b00663 \\
acs.orglett.5b00740 & acs.orglett.5b00776 & acs.orglett.5b00805 & acs.orglett.5b01044 \\
acs.orglett.5b01077 & acs.orglett.5b01309 & acs.orglett.5b01385 & acs.orglett.5b01692 \\
acs.orglett.5b01754 & acs.orglett.5b01842 & acs.orglett.5b01872 & acs.orglett.5b02003 \\
acs.orglett.5b02279 & acs.orglett.5b02498 & acs.orglett.5b02545 & acs.orglett.5b02680 \\
acs.orglett.5b02709 & acs.orglett.5b02743 & acs.orglett.5b03104 & acs.orglett.5b03589 \\
acs.orglett.5b03590 & acs.orglett.6b00233 & acs.orglett.6b00326 & acs.orglett.6b00661 \\
acs.orglett.6b01059 & acs.orglett.6b01181 & ja001164i & ja0014685 \\
ja0056062 & ja011003u & ja012253d & ja012741l \\
ja0161958 & ja016495p & ja0171299 & ja017617g \\
ja0176346 & ja026640e & ja026703t & ja0289088 \\
ja029499i & ja030125e & ja030261j & ja040054z \\
ja042849b & ja0516864 & ja052327b & ja053368a \\
ja053650h & ja054378e & ja0547477 & ja0551382 \\
ja060064v & ja063878k & ja064212t & ja065718e \\
ja074044k & ja075824w & ja076333e & ja1048847 \\
ja106807u & ja1078199 & ja107927b & ja200818w \\
ja2014746 & ja2031294 & ja204366b & ja206047h \\
ja2070522 & ja207331m & ja208286b & ja211778j \\
ja300396h & ja3058138 & ja3066978 & ja307151x \\
ja312277g & ja402810t & ja406383h & ja407179c \\
ja407689a & ja408031s & ja408733f & ja410533y \\
ja501560x & ja5017206 & ja5080739 & ja511335v \\
ja511913h & ja801487v & ja806060a & ja806814c \\
ja900722q & ja9039289 & ja905415r & ja953272o \\
ja9535975 & ja954050t & ja960062i & ja9612413 \\
ja974106e & ja980022+ & ja9810742 & ja983111v \\
ja991729e & ja992608h & jacs.5b00936 & jacs.5b05415 \\
jacs.5b05596 & jacs.5b05792 & jacs.5b07904 & jacs.5b11315 \\
jacs.5b12989 & jacs.6b00143 & jacs.6b01306 & jo000081h \\
jo000585f & jo000694u & jo000745n & jo0007837 \\
jo001223a & jo001386z & jo0014156 & jo0014414 \\
jo0014820 & jo001614p & jo001700p & jo0056489 \\
jo010170+ & jo010230b & jo010297z & jo010404p \\
jo0108865 & jo010904i & jo0109321 & jo011082s \\
jo015508e & jo0157425 & jo015897c & jo025690z \\
jo025987x & jo0524728 & jo102193q & jo2001275 \\
jo2001534 & jo2003264 & jo200480h & jo200666z \\
jo2008675 & jo200877k & jo200882k & jo201056f \\
jo201098c & jo201478d & jo201489z & jo201975b \\
jo201996w & jo2020856 & jo202294k & jo202324f \\
jo202571h & jo300771f & jo302142v & jo302288z \\
jo302439t & jo4004426 & jo400501k & jo4007046 \\
jo400755q & jo401195c & jo4014707 & jo401608v \\
jo402399n & jo4026034 & jo4027148 & jo402749d \\
jo402763m & jo500412w & jo500696n & jo501006u \\
jo501014e & jo501180a & jo501216h & jo501736w \\
jo501785d & jo501910q & jo501913z & jo502408z \\
jo502578x & jo5026145 & jo502752u & jo800904u \\
jo951894m & jo951899j & jo9519672 & jo9521209 \\
jo9602433 & jo960401q & jo960838y & jo961049j \\
jo961323+ & jo961365y & jo961824v & jo962200s \\
jo970671o & jo9708497 & jo971595s & jo9716338 \\
jo9717245 & jo980058k & jo980181b & jo980755c \\
jo980767y & jo981125d & jo981397g & jo9816515 \\
jo982004g & jo982024i & jo9901541 & jo9902998 \\
jo990528q & jo9906328 & jo990938e & jo991071n \\
jo9911286 & jo991198c & jo991283k & jo991457y \\
jo991524o & jo991681n & jo991700t & jo9919409 \\
ol0002368 & ol006041h & ol006129v & ol006192k \\
ol006383n & ol006614q & ol0069002 & ol010283f \\
ol0157003 & ol015948s & ol016212y & ol016466j \\
ol016689+ & ol016693l & ol0171867 & ol0173127 \\
ol025887d & ol026156g & ol026509b & ol027494k \\
ol034434l & ol034469l & ol0348957 & ol0349920 \\
ol035127i & ol035681s & ol036111v & ol0361507 \\
ol0362663 & ol036510q & ol047761h & ol0480731 \\
ol048585f & ol0487783 & ol048861q & ol049640n \\
ol050019c & ol050791f & ol051245p & ol051342i \\
ol051365x & ol0513995 & ol0514606 & ol051488h \\
ol051901l & ol051920v & ol052113z & ol052245s \\
ol052474e & ol053021c & ol0600584 & ol060123+ \\
ol060246u & ol0604623 & ol060473w & ol060531d \\
ol060664z & ol060868f & ol0610183 & ol061289d \\
ol0616236 & ol0619157 & ol0701619 & ol070339r \\
ol0703579 & ol071385u & ol071386m & ol100073y \\
ol100734t & ol1009703 & ol101406k & ol101839m \\
ol1018773 & ol1022036 & ol102738b & ol102784c \\
ol1030487 & ol200038n & ol200288w & ol200703g \\
ol200717n & ol200849k & ol201201a & ol2017438 \\
ol2017998 & ol202027k & ol202381m & ol202395s \\
ol202499g & ol202528k & ol203001w & ol300353w \\
ol300387f & ol300808c & ol300842d & ol301114z \\
ol301535j & ol301556a & ol301852m & ol301863j \\
ol3023177 & ol3023903 & ol302400p & ol302668y \\
ol302863r & ol302997q & ol303154k & ol303452r \\
ol303482k & ol400025a & ol400110c & ol4005905 \\
ol401042b & ol401179k & ol4012794 & ol4013926 \\
ol401443a & ol401535k & ol401571r & ol4017244 \\
ol4017854 & ol401812h & ol401881n & ol401909z \\
ol402047d & ol402138y & ol4025293 & ol402871f \\
ol402981z & ol403122a & ol5000692 & ol500121z \\
ol500248h & ol500444z & ol500618w & ol5007604 \\
ol501019y & ol501085y & ol501165h & ol5012407 \\
ol501422k & ol501424f & ol501514b & ol5020043 \\
ol502423k & ol502425f & ol502664f & ol502681y \\
ol5028392 & ol502842f & ol5029892 & ol502998n \\
ol503404p & ol503587n & ol503618m & ol503708v \\
ol7014434 & ol701624y & ol7019636 & ol702044z \\
ol7028367 & ol800086s & ol8001706 & ol800288b \\
ol800418m & ol8005198 & ol800523j & ol800527p \\
ol8006106 & ol8006259 & ol801034x & ol801035c \\
ol801163v & ol8013717 & ol801498u & ol801788t \\
ol801791g & ol8019605 & ol802005n & ol802073q \\
ol802141g & ol802297h & ol802556f & ol802669r \\
ol802674r & ol9005079 & ol9005322 & ol901584g \\
ol901684h & ol901760a & ol990836a & ol9909583 \\
ol9913542 & ol991356m & op000061h & op000070q \\
op0000879 & op000095p & op000111i & op000298d \\
op010052o & op010068e & op0100706 & op010073i \\
op010097p & op0101106 & op0102013 & op010232y \\
op0155211 & op020010f & op020019h & op020049k \\
op020098x & op020211j & op0202179 & op025501e \\
op025538z & op0255478 & op0255736 & op025610t \\
op0256183 & op025619v & op0300488 & op030202q \\
op034033l & op034064g & op0340661 & op0340816 \\
op0340964 & op034181b & op034198u & op049803n \\
op049889k & op049899l & op049953y & op049954q \\
op050040t & op050061n & op050077d & op050087e \\
op0501242 & op050151s & op0501803 & op050182n \\
op050193g & op0600106 & op060099f & op060114g \\
op060118l & op0601316 & op060155c & op0601619 \\
op060175e & op0602270 & op060249m & op100010n \\
op100072y & op100103v & op100104z & op100108j \\
op100113j & op100197g & op100202j & op100205s \\
op100210s & op100267p & op100335q & op200005e \\
op2000089 & op200011x & op200019k & op200038y \\
op200052z & op200086t & op2001047 & op200112g \\
op200174k & op200176f & op2001832 & op200234j \\
op200312m & op200313v & op2003216 & op200334x \\
op200351g & op3000042 & op300031r & op300058f \\
op300059b & op300087r & op300101d & op3001355 \\
op300147f & op300162d & op300170q & op300171m \\
op3001788 & op300181r & op300205j & op300209p \\
op300213s & op300216x & op300235t & op300252n \\
op3002883 & op3003097 & op300331b & op300341n \\
op300343q & op300363s & op300364p & op400050n \\
op400055z & op400113a & op400135y & op400242j \\
op400269b & op400278d & op400292m & op4003467 \\
op500072b & op500102h & op5001226 & op5001385 \\
op5001463 & op500221s & op500224x & op500234a \\
op5002462 & op500250b & op5003165 & op500334b \\
op700009t & op7000172 & op700026n & op700039r \\
op700060e & op7001485 & op700160a & op7001694 \\
op700175d & op700178q & op7001886 & op700249f \\
op700253t & op700274v & op7002826 & op700292s \\
op800033c & op8000756 & op800091p & op800136f \\
op8001596 & op800177x & op8001799 & op800189g \\
op8002097 & op8002486 & op800270e & op900008a \\
op900056s & op9000687 & op900102a & op9001824 \\
op900188v & op900197r & op900242x & op9002533 \\
op9002642 & op900265h & op960008m & op9600419 \\
op970105v & op970113b & op9701245 & op980039c \\
op9800717 & op980075b & op980079g & op980184q \\
op9802071 & op990044w & op990049t & op990050s \\
op990067a & op990099y \\
\end{longtable}
}

\bibliography{reference}